\documentclass[conference]{IEEEtran}
\IEEEoverridecommandlockouts
\usepackage{cite}
\usepackage{amsmath,  amssymb,  amsfonts}
\usepackage{algorithmic}
\usepackage{graphicx}
\usepackage{textcomp}
\usepackage{multirow}
\usepackage{xcolor}
\usepackage{colortbl}
\usepackage{graphicx}
\usepackage{balance}

\usepackage[top=60pt, bottom=54pt, left=54pt, right=54pt]{geometry}
\def\BibTeX{{\rm B\kern-.05em{\sc i\kern-.025em b}\kern-.08em
    T\kern-.1667em\lower.7ex\hbox{E}\kern-.125emX}}
\begin{document}

\title{AHPPEBot: Autonomous Robot for Tomato Harvesting based on Phenotyping and Pose Estimation
\thanks{
\vspace{0.0em}
\textsuperscript{*}Work done during an internship at Beijing AIForce Technology.

This work is supported by the National Natural Science Foundation of China (No. 62371013), Beijing AIForce Technology, the Beijing Natural Science Foundation (No. 4222025), QIYUAN LAB Innovation Foundation (Innovation Research) Project (No. S20210201107).

\textsuperscript{1}Faculty of Information Technology, Beijing University of Technology, Beijing 100124, China. lixingxu@emails.bjut.edu.cn, hanyiheng@bjut.edu.cn

\textsuperscript{2}Beijing AIForce Technology Co., Ltd., 6 Chuangye Road, Beijing 100085, China. \{yangshun, zhengsiyi\}@aiforcetech.com

\textsuperscript{†} Corresponding author: Nan Ma. manan123@bjut.edu.cn

\vspace{-1.0em}
}
}

\author{Xingxu Li\textsuperscript{1,2*},
        Nan Ma\textsuperscript{1†},
        Yiheng Han\textsuperscript{1},
        Shun Yang\textsuperscript{2},
        and Siyi Zheng\textsuperscript{2},
        }

\maketitle

\begin{abstract}
To address the limitations inherent to conventional automated harvesting robots specifically their suboptimal success rates and risk of crop damage, we design a novel bot named AHPPEBot which is capable of autonomous harvesting based on crop phenotyping and pose estimation. Specifically, In phenotyping, the detection, association, and maturity estimation of tomato trusses and individual fruits are accomplished through a multi-task YOLOv5 model coupled with a detection-based adaptive DBScan clustering algorithm. In pose estimation, we employ a deep learning model to predict seven semantic keypoints on the pedicel. These keypoints assist in the robot's path planning, minimize target contact, and facilitate the use of our specialized end effector for harvesting. In autonomous tomato harvesting experiments conducted in commercial greenhouses, our proposed robot achieved a harvesting success rate of 86.67\%, with an average successful harvest time of 32.46 s, showcasing its continuous and robust harvesting capabilities. The result underscores the potential of harvesting robots to bridge the labor gap in agriculture.
\end{abstract}

\begin{IEEEkeywords}
Precision farming; Selective harvesting; Agricultural robotics; Plant phenotyping; Pose estimation
\end{IEEEkeywords}

\section{Introduction}
Labor shortages in agricultural production limit the expansion of production scales[1]. Implementing automation systems is a feasible solution to enhance productivity[2]. Research on robotic harvesting has garnered increasing attention, and substantial progress has been achieved in crops such as strawberries[3], apples[4], bell peppers[5], and tomatoes[6]. The widely adopted perception methods and harvesting procedures can be summarized as follows: maturity is categorized into a binary classification using supervised learning, where maturity levels are directly mapped to harvestable and non-harvestable states. Once the detection is completed, point clouds of the target are acquired through segmentation models, and these clouds are then fit into geometric shapes like spheres or cylinders, facilitating the determination of crop posture and planning of the appropriate grabbing angle.

\begin{figure}[tbp]
\centerline{\includegraphics[width=1.0\linewidth]{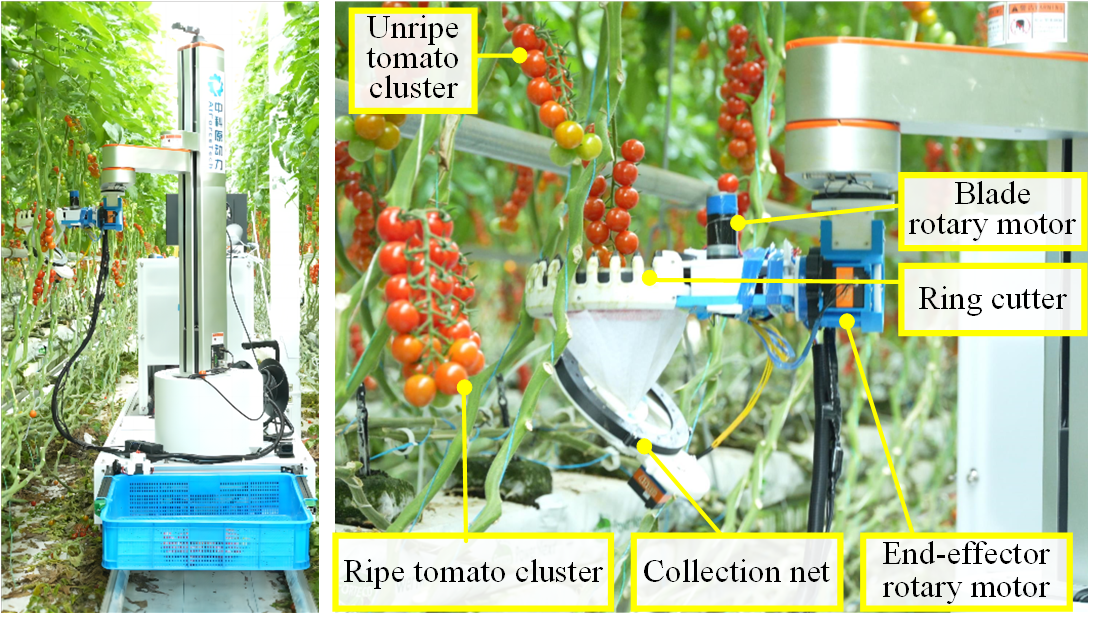}}
\vspace{-1.0em}
\caption{AHPPEBot Overview: The robot perceives nearby tomato trusses and autonomously makes decisions based on various information such as arm workspace, ripeness, and pose, ultimately using a circular cutter to harvest the target.}
\label{fig}
\vspace{-1.0em}
\end{figure}

While these traditional methods have been applied to many crops, they have limitations regarding the automated harvesting of truss tomatoes. Firstly, as truss tomatoes are a fresh consumption economic crop, their maturity requires a more detailed assessment to ensure edible quality. The interpretability and adjustability of the binary classification method based on deep learning are insufficient, hampering the robot's selective harvesting and quality management of the harvested product. Secondly, unlike strawberries or citrus fruits, truss tomatoes cannot be harvested by grabbing and pulling, as this may lead to crop or facility damage. The ideal harvesting method is cutting the peduncle, but due to its slender nature and color similarity to the background, detection and cutting point localization present challenges.

This paper designs AHPPEBot, a robot Autonomous Harvesting tomato trusses based on Phenotyping and Pose Estimation. To improve the robot's success rate, efficiency, and safety, we have made advancements in three crucial areas:

\begin{enumerate}[]

\item Phenotyping: We propose an adaptive DBScan clustering algorithm predicated on object detection, designed to associate individual fruits with their corresponding tomato trusses, facilitating the extraction of pertinent information, encompassing the overall maturity level of the tomato truss, the number of fruits, and their respective volumes.

\item Pose estimation:  We introduce a deep learning-based keypoint detection method for pose estimation of tomato trusses, enabling arm path planning and movement with reduced contact with the target.

\item  Harvesting System: Integrating phenotypic and postural information, the robot achieves full-state perception of the tomato truss, autonomously selecting optimal harvest targets. Leveraging posture key points, it strategically plans the arm trajectory, complemented by a novel designed end-effector for precise harvesting.
\end{enumerate}

\section{Related Work}
Recognizing, parsing, and understanding crops in the environment is foundational to enabling robots to perform selective harvesting autonomously. Early research primarily identified and located fruits through image processing techniques such as color segmentation and morphological operations or through machine learning approaches[7-10, 13]. Recent advancements have shown that perception techniques based on deep learning models exhibit greater robustness in detection, classification, and segmentation challenges in real-world scenarios compared to traditional methods. Zhang et al.[11] utilized CNNs for the maturity classification of tomato images, while Afonso et al.[12] employed Mask R-CNN to detect ripe and unripe tomatoes in greenhouses, achieving commendable accuracy and robustness. 

The precise orientation or posture of crops is imperative for accurate harvesting. Wei et al.[14] extracted grape bunch pixel regions using Mask R-CNN and derived corresponding point clouds. These point clouds were fitted into cylindrical shapes using the RANSAC algorithm to determine posture. Rong et al.[15] utilized YOLOv4-Tiny to detect the main body and stalk of tomato bunches. After semantic segmentation of the stalk with the YOLACT++ model, they used least squares to fit it into a curve, obtaining three key points, and subsequently estimated the stalk's posture through a geometric model. Li et al.[16] employed a multi-camera system and a DCNN model to detect apple targets, reconstructed obscured parts from the visible target point cloud, and the estimated fruit center. In Alessandra et al.'s [17] method, Five key points are defined for each tomato, serving as references for the end-effector performing the grasping action. Fan et al.[18] proposed a keypoint-based method for estimating the pose of tomato trusses. Their approach is limited to trusses containing only six fruits.

Even when crop detection and harvesting point localization are accomplished, the actual harvesting process can still encounter failures. During the act of harvesting, robotic arms equipped with end-effectors may fail due to obstacles, pose estimation errors, or damage to the crop. In 2017, Bac et al.[19] designed a bell pepper harvesting robot under the CROPS (the Clever Robots for Crops) project, conducting experiments in a commercial greenhouse using two different end-effectors in untrimmed and trimmed environments. Lehner et al.[9] designed a bell pepper harvesting robot named "Harvey," which harvests by suctioning the target and cutting the peduncle with a blade. In particular failure cases, the irregular morphology of the bell peppers led to deviations in the model’s estimated grasping angle. Consequently, the blade failed to sever the pedicel accurately, or inadvertently caused damage to the fruit. Gao et al.[6] investigated a two-fingered gripper end-effector that harvests individual fruits from cherry tomato trusses. 

Reviewing prior research, the reasons for harvesting failures can be categorized as follows: inaccurate fruit localization, obstructions from leaves or fruit clusters, extreme fruit positions, crop damage, and separation failures.

\section{Method}

AHPPEBot is designed for automated truss tomato harvesting in commercial greenhouse environments, as shown in Figure 1. The harvesting system can be divided into four main components: "Phenotyping," "Pose Estimation", "Fusion of Information," and "Decision and Motion Planning". The flowchart of the system algorithm and workflow is illustrated in Figure 2. We anticipate that by optimizing perception techniques and end-effector design and precisely planning the robotic arm's motion based on crop posture, we can minimize unnecessary contact, thereby achieving our goal of enhancing the harvesting success rate and safety.

\begin{figure*}[tb]
\centerline{\includegraphics[width=1.0\linewidth]{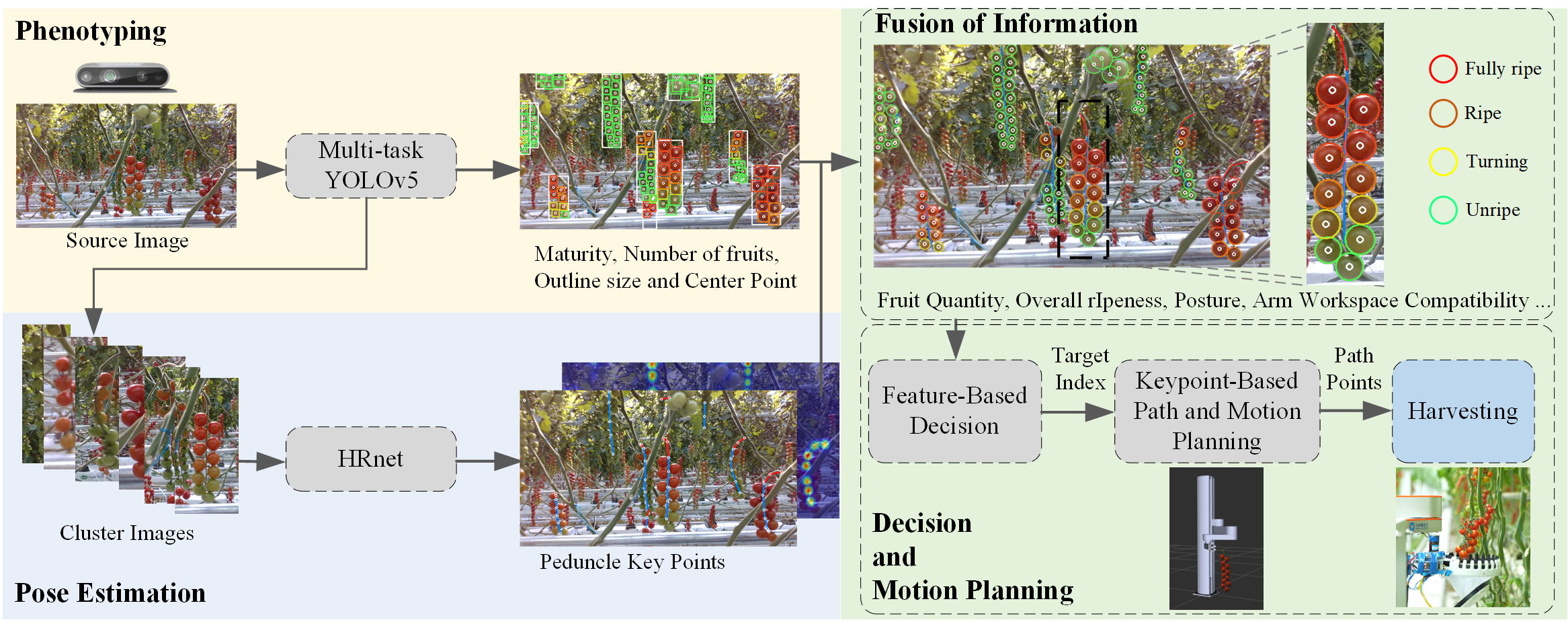}}
\vspace{-1.0em}
\caption{Automated Harvesting System Workflow: During the phenotyping, the number and ripeness of each tomato truss are determined (different color prediction frames indicate varying maturity levels). The pose detection captures the key points of the tomato trusses. By integrating and encoding both phenotypic and pose information, a target is selected. Subsequently, based on its pose, the robotic arm's path is planned, and harvesting is executed.}
\label{fig}
\vspace{-1.0em}
\end{figure*}

\subsection{Platform Design}

The hardware architecture of AHPPEBot comprises five main components: two RGB-D cameras, a robotic arm, an end-effector, a computing unit, and a mobility chassis. In consideration of the narrow space within greenhouses, we opted for a SCARA robotic arm, augmented with a rotary motor at its end, as depicted in Figure 1. This configuration serves to simplify path planning for the robotic arm while maintaining flexibility.

The exterior of the effector features multiple spike-shaped guiding grooves and a saw blade. The groove size is tailored based on the width of the tomato vine and truss peduncle, ensuring that only the peduncle can fit and be cut while preventing the vine from entering, enhancing safety. Once the peduncle is severed, the tomato truss temporarily collects in a mesh pocket. 

\subsection{Phenotyping}
Maturity level, fruit count, and fruit size are important phenotypic information for harvesting and quality grading. We adopt a method that infers the overall maturity and quality grading of the truss based on the status of each fruit. In this paper, fruit maturity is categorized into four stages based on agronomic standards: green mature, turning, ripe, and fully ripe, as illustrated in Figure 2. Concurrently, we stipulate that if the terminal fruit on a truss reaches at least the turning stage and the rest of the fruits are ripe or beyond, the truss is considered mature overall and suitable for harvesting, storage, and transport. This criterion aligns with the harvesting standards of the greenhouse production base where our experiments took place.

We enhanced the YOLOv5 model for multi-tasking, introducing an additional branch in the prediction section to determine fruit maturity. The loss function for our multi-task YOLOv5 is designed based on the original YOLOv5's loss function:

\begin{equation}
\mathcal{L}=\lambda_1 \mathcal{L}_{{cls}}+\lambda_2 \mathcal{L}_{{conf}}+\lambda_3 \mathcal{L}_{{box}}+\lambda_4 \mathcal{L}_{{rip}}
\end{equation}
where $\lambda$ represents the weight of different loss functions. $\mathcal{L}_{{cls}}$is the classification loss, $\mathcal{L}_{{cls}}$ is the confidence loss and $\mathcal{L}_{{box}}$ represents the bounding box loss. The binary cross-entropy loss function $\mathcal{L}_{{rip}}$ is employed to compute the tomato fruit maturity loss.

While the model can detect each individual fruit, it does not provide the relationship between the fruits and the tomato trusses. We match and group fruits based on their 2D/3D spatial relationships with the tomato trusses, thus obtaining information on the quantity and maturity of fruits within a given truss. In the 2D space, we calculate the Intersection over Union (IOU) between the bounding boxes of the fruits and the tomato trusses to determine if a fruit belongs to a specific truss. However, when multiple tomato trusses overlap in the camera's field of view, this algorithm may produce matching errors. To address this issue, we employ point cloud clustering to distinguish overlapping tomato trusses in the foreground and background, thereby determining the ownership of fruits.

Traditional DBScan clustering suffers from extensive and inefficient nearest neighbor searches and random initial point selection, affecting its computational speed. We propose an adaptive DBScan algorithm based on object detection. The bounding boxes in the detection results provide prior knowledge of the spatial positions of the targets. Precise cropping of the input point cloud is achieved using the exact bounding boxes of the tomato trusses, eliminating the need for full point cloud computation. The center points of the fruit bounding boxes are used as the initial point set for clustering, enabling adaptive selection of high-density target point clouds. By employing adaptive input cropping and initial point selection, the number of nearby searches can be significantly reduced, leading to a notable improvement in computational speed. Additionally, the threshold values of the clustering algorithm can be determined based on the structural prior knowledge of various tomato truss varieties.

To swiftly estimate the volume of fruits and determine their spatial positions, we do not partition the precise point cloud of each fruit from the point cloud returned by the depth camera. Instead, we approximate the contour of the fruits in the images using a circle with a radius equal to the average width and height of the rectangular predicted bounding box. Combined with the intrinsic parameters of the depth camera, we perform back-projection to generate a spherical virtual point cloud in three-dimensional space, preparing for collision detection during path planning, as illustrated in Figure 2. This estimation method has low computational complexity and errors are within an acceptable range for our harvesting approach.

\subsection{Pose Estimation}

To better support robotic arm path planning, we incorporated structural prior knowledge of tomato trusses and defined seven peduncle keypoints. Their nomenclature and abbreviations are shown in Figure 3. SP, CP, and FP along with their connecting curves represent segments of the peduncle that can be cut. SP is situated at the junction of the tomato truss peduncle and the main stem, serving as both the starting point of the tomato curve and the anticipated cutting position for our end-effector. CP is located at the maximum curvature of the peduncle and is also the traditional cutting point for clamp-style end-effectors. Ideally, in our robot, cutting should occur close to the SP position of the peduncle. FP is where the first fruit petiole connects to the peduncle, while EP is at the end of the peduncle, and in combination with other peduncle and fruit keypoints, indicates the posture of the tomato truss. This area is to be avoided contact or cutting.

Even when annotating peduncle keypoints for the same tomato truss, different annotators may produce results that are not entirely consistent, potentially resulting in varying degrees of discrepancy. For instance, the positions of QP, MP, and TQP are evenly distributed along the peduncle area from FP to EP. Their structural features are less prominent compared to SP and CP. Similar to the COCO human keypoint dataset [20], we use the OKS (Object Keypoint Similarity) value as the evaluation metric for peduncle keypoint prediction accuracy. The 'sigmas' parameter for peduncle keypoints is obtained by calculating the standard deviation between multiple human annotator keypoints and expert ground truth annotations.

The difference lies in the fact that not all keypoints in the COCO human keypoint dataset are equally important. For harvesting tasks, the importance of each keypoint and the consequences of prediction errors vary. SP, CP, and FP peduncle keypoints require the highest prediction and localization accuracy possible. If their predicted positions do not align perfectly with the peduncle, harvesting is likely to fail, and there may even be damage to the tomato vines. The other points are primarily used to avoid collisions between the end-effector and tomato trusses and do not require ultra-high precision. The OKS calculates the overall similarity between the ground truth and predicted values of multiple keypoints on the target. One potential issue is that a model with a high accuracy evaluation score may perform well in predicting keypoints from FP to EP, but may lack precision in predicting SP and CP. This contradicts our harvesting objectives.

We manually fine-tuned the' sigmas' values obtained for each keypoint to obtain a keypoint detection model based on OKS evaluation that better aligns with harvesting requirements. The manual adjustment involved reducing the 'sigmas' values for SP and CP points while increasing the 'sigmas' values for the remaining points. This adjustment signifies our intention for the model evaluation to lean towards higher accuracy prediction of SP and CP keypoints.

For detecting keypoints on tomato peduncles, we employed the HRnet-w48 model. Additionally, several state-of-the-art networks on the COCO keypoint detection dataset were experimented with, including RTMPose, CID, UDP, and DarkPose, with CID, DarkPose, and UDP all using HRnet as their backbone. Moreover, we explored regression-based keypoint detection approaches, such as YOLOv8-pose, and models using Resnet101 as the backbone for regression prediction. 

\begin{figure*}[tb]
\centerline{\includegraphics[width=1.0\linewidth]{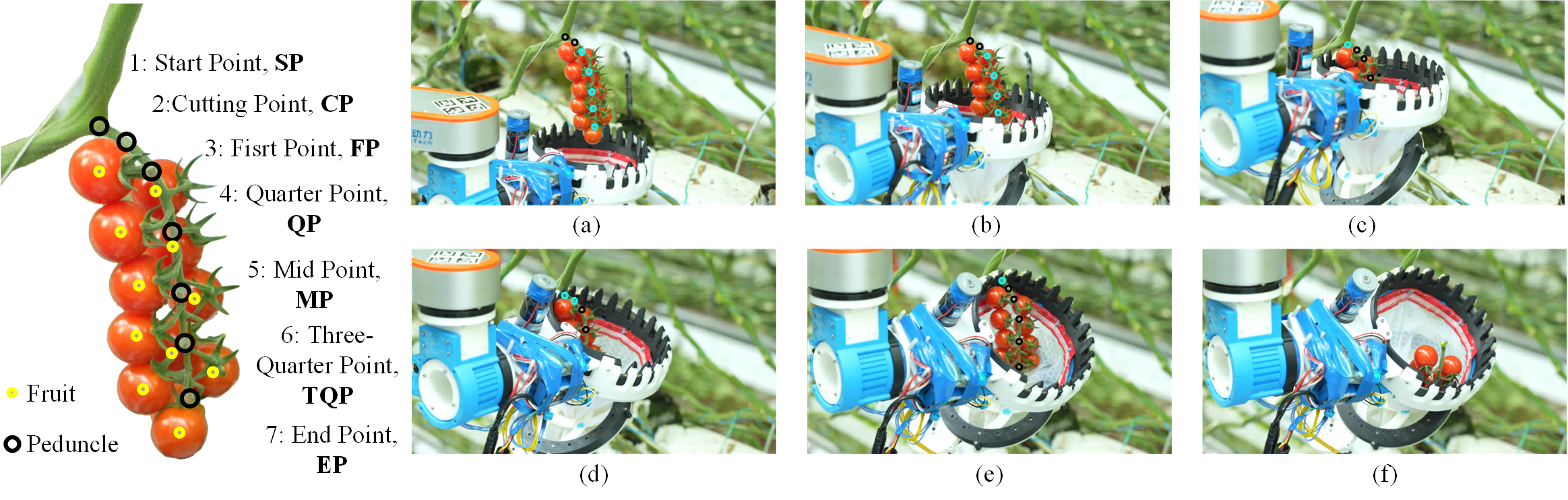}}
\vspace{-1.0em}
\caption{A tomato truss is represented by a combination of 7 peduncle keypoints and multiple fruit keypoints. The harvesting process, termed "bottom-up wrapping," involves (a) initial positioning below the target; (b) slow ascent and translation along the peduncle curve to avoid collisions with the target until all fruits are fully enveloped; (c) bringing the edge of the end-effector close to the SP point; (d) rotating the end-effector, causing the peduncle to fall into the slot and be cut by the blade; (e) applying continued pressure through rotation; (f) the peduncle is severed and falls into the collection mesh. Key keypoints predominantly utilized in each stage are denoted in blue.}
\label{fig}
\vspace{-1.0em}
\end{figure*}

\subsection{Decision and Motion Planning}

In the greenhouse, limited operational space and entangled vines challenge automated harvesting tasks. We contend that the operation of harvesting robots must be predicated upon the principle of "do no harm" to the surroundings. Accordingly, we have designed and optimized decision-making processes with "minimizing risks whenever possible" as a central tenet.

By integrating the local information of the tomato trusses obtained from the Phenotyping phase with the holistic posture information acquired during the Pose estimation phase, we encode the features of each tomato trusses object. Utilizing this feature data, we further refine the selection of optimal harvesting targets. Targets that are either immature or of subpar quality are excluded. tomato trusses with extreme growth positions or orientations are also discarded. For instance, when harvesting tomato trusses facing the interior of cultivation troughs or positioned opposite to the direction of vine suspension, there is a high likelihood of occurrences such as mechanical arm collisions with heating pipes or entanglement with vines. Excluding such targets would effectively reduce the risk of accidents and contribute to enhancing the overall continuity and success rate of the harvesting process.

Upon target selection, the robot engages in end-effector path planning based on target volume and pose keypoints, refer to Figure 3. Initially, the EP, TQP, MP, QP, and FP points serve as end-effector motion pathpoints, generating a smooth end-effector trajectory. To enhance harvesting success, collision detection is performed to ensure that the inner wall and upper edge of the end-effector do not collide with the tomato trusses. Leveraging prior knowledge of tomato truss growth, it is observed that fruits predominantly grow towards the normal direction of the peduncle curve. If there is a risk of collision, the end-effector path points are shifted outward along the normal direction of the peduncle curve. Subsequently, the end-effector follows the planned trajectory to envelop the main body of the tomato truss containing the fruits, with the edge containing the blade reaching the SP point. The end-effector rotates, applying pressure and cutting the peduncle, thereby completing the harvesting process.

\section{Experiment Results}

\subsection{Dataset and Training}
Due to the absence of publicly available cherry tomato datasets, we collected image data from a commercial greenhouse in Haidian District, Beijing. Two datasets were generated after manual annotation by experts: Dataset-1, tailored for tomato object detection and ripeness differentiation, comprises 2,000 annotated images with 112,000 targets. Dataset-2, designed for peduncle keypoint detection, encompasses 1,000 images with 5,432 annotated subjects. For phenotypic multi-task modeling, we employed the Yolov5m model and trained it on Dataset-1, designating 300 images for validation. Multiple models destined for peduncle keypoint prediction were trained using Dataset-2, allocating 150 images for the validation set.

\subsection{Phenotyping Test}

Our trained multi-task YOLOv5m achieved an mAP of 90.18\% on the validation set for tomato trusses and four maturity stages of fruits with IOU threshold set at 0.5. Our assessment of overall ripeness and other phenotypic applications is based on this model and the adaptive DBscan method. Subsequently, through an enhanced adaptive clustering algorithm, fruits were systematically matched with their corresponding trusses, facilitating an inference of the aggregate ripeness from individual fruit metrics. As a benchmark, the conventional method utilized a supervised binary classification model to evaluate the truss's maturity directly. Within a validation dataset consisting of 300 image samples, both methodologies were rigorously examined, and the findings are delineated in Table 1.

\begin{table}[htbp]
\centering
\vspace{-0.5em}
\fontsize{8}{10.8}\selectfont
\caption{Accuracy Comparison of tomato trusses Maturity Assessment on the Validation Set: Our Approach (Phenotyping based on Object Detection) vs. Traditional Methods (Supervised Deep Learning Classification Models).}
\label{tab:my-table}
\begin{tabular}{cc} 
\hline
\textbf{Method} & \textbf{Result} \\ 
\hline
\textbf{Our method} & \textbf{0.8971} \\
Mobilenet-v2 & 0.8604 \\
Resnet18 & 0.8101 \\
Efficientnet-b4 & 0.8449 \\
Swin-Transformer small & 0.8527 \\
\hline
\end{tabular}
\vspace{-0.5em}
\end{table}

The experimental results demonstrate that our method exhibits a significant accuracy advantage in estimating the maturity of entire tomato trusses. Moreover, this method aligns well with agronomic standards, maintaining excellent interpretability and adjustability. To illustrate, if there's a shift in ripeness criteria to consider a tomato truss as mature only when every individual fruit achieves or surpasses a certain maturity threshold, the conventional classification approach would demand an exhaustive re-evaluation and adjustment of annotations across the dataset.

\subsection{Pose estimation test}

We compiled a pedicel keypoint validation set comprising 150 images to evaluate the pose estimation accuracy of various models. The test results are presented in Table 2.

\begin{table}[htbp]
\centering
\vspace{-0.5em}
\fontsize{8}{10.8}\selectfont
\caption{Performance of Various Models on the Pedicel Keypoint Validation Set using OKS Metric at a Threshold of 0.75.}
\label{tab:my-table}
\begin{tabular}{cccc} 
\hline
\textbf{Category} & \textbf{Model} & \textbf{Input size} & \textbf{OKS@.75} \\ 
\hline
\multirow{6}{*}{\begin{tabular}[c]{@{}c@{}}Heatmap \\ based\end{tabular}} & \textbf{Our method} & 256x192 & \textbf{0.7561} \\
 & HRnet\_w32 & 256x192 & 0.6968 \\
 & DarkPose & 384x288 & 0.7113 \\
 & CID & 512x512 & 0.4895 \\
 & UDP & 384x288 & 0.7456 \\
 & RTMPose\_L & 256x192 & 0.7452 \\
\multirow{2}{*}{\begin{tabular}[c]{@{}c@{}}Regression \\ based\end{tabular}} & Yolov8-pose & 640x640 & 0.5307 \\
 & RES101\_RLE & 256x192 & 0.5447 \\
\hline
\end{tabular}
\vspace{-0.5em}
\end{table}

Based on the experimental results, when using the OKS metric with a threshold of 0.75, our method exhibited outstanding performance, achieving a precision of 0.7561. Although our algorithm lags in inference speed, considering that the operational efficiency of the harvesting robot is predominantly constrained by the several-second action cycle of its mechanical components, this shortfall seems relatively insignificant compared to perception speed.

Furthermore, we observed that keypoint prediction models based on heatmaps generally outperform models that directly regress the keypoint positions in detecting peduncle keypoints. This is likely related to the small pixel area that the peduncle occupies in the image. For instance, in a 720P frame containing a tomato truss image with dimensions 138x438 pixels, the pedicel has a pixel width of 8. Prior to input into the HRnet with an input layer of 192x168, this pedicel pixel region is magnified. Conversely, models employing global computations, such as YOLO with an input layer of 640x640, would compress the pedicel image, thereby compromising precise prediction.

\subsection{Automated Harvesting experiment}
In a greenhouse, two rounds of experiments were conducted. First, under a controlled scenario, we evaluated the robot's performance using a harvesting path planning based on pose estimation, also known as 'Bottom-up wrapping', and its end-effector's proficiency when confronting tomato trusses of varied orientations. Second, we tested AHPPEBot's continuous harvesting capability in an unaltered environment.

\subsubsection{Harvesting Test Under Different Target Poses}
In the experiment, we manually pruned to ensure only one cluster of tomatoes was present in the robot's workspace, preventing interference with the vines. Upon receiving a start command, the robot initiated the harvesting. If three consecutive attempts failed, or if there were safety concerns or damage to the tomato, the action was deemed unsuccessful. In such cases, the operator intervened, halting the operation and removing the target tomato trusses. Upon successful harvesting, the robot moved to the next pick point.

In preliminary tests, due to constraints from planting infrastructure and the direction of tomato plant hangings, clusters with orientations facing left or inwards (growing towards the cultivation trough) were challenging to harvest from the robot's viewpoint. Consequently, these outliers were excluded, focusing only on clusters suitable for harvesting, those facing right or towards the robot. The success rates for critical steps throughout the harvesting process are detailed in Table 3.

\begin{table}[htbp]
\centering
\fontsize{8}{10.8}\selectfont
\vspace{-0.5em}
\caption{tomato trusses Harvesting Success Rates for Two Poses Under Manual Intervention: SP Point Prediction (Accuracy in Locating Points on the Pedicel), Bottom-up Wrapping (Success in Target Encasement), Detachment Efficacy (Successful Pedicel Cutting), and Overall Harvesting Integrity (Preservation and Correct Maturity of tomato trusses).}
\begin{tabular}{lllll} 
\hline
\begin{tabular}[c]{@{}l@{}}\textbf{Tomato}\\\textbf{ Bunch }\\\textbf{ Pose}\end{tabular} & \begin{tabular}[c]{@{}l@{}}\textbf{SP Point }\\\textbf{ Identification}\end{tabular} & \begin{tabular}[c]{@{}l@{}}\textbf{Bottom-up}\\\textbf{ Wrapping}\end{tabular} & \textbf{Detach} & \textbf{Harvesting} \\ 
\hline
Front & \begin{tabular}[c]{@{}l@{}}86.36\%\\ (19/22)\end{tabular} & \begin{tabular}[c]{@{}l@{}}95.45\%\\ (21/22)\end{tabular} & \begin{tabular}[c]{@{}l@{}}95.24\%\\ (20/21)\end{tabular} & \begin{tabular}[c]{@{}l@{}}90.90\%\\ (20/22)\end{tabular} \\
Right & \begin{tabular}[c]{@{}l@{}}84.62\%\\ (22/26)\end{tabular} & \begin{tabular}[c]{@{}l@{}}92.31\%\\ (24/26)\end{tabular} & \begin{tabular}[c]{@{}l@{}}95.83\%\\ (23/24)\end{tabular} & \begin{tabular}[c]{@{}l@{}}88.46\%\\ (23/26)\end{tabular} \\
 & \begin{tabular}[c]{@{}l@{}}85.41\%\\ (41/48)\end{tabular} & \begin{tabular}[c]{@{}l@{}}93.75\%\\ (44/48)\end{tabular} & \begin{tabular}[c]{@{}l@{}}95.56\%\\ (43/45)\end{tabular} & \begin{tabular}[c]{@{}l@{}}89.58\%\\ (43/48)\end{tabular} \\
\hline
\end{tabular}
\vspace{-0.5em}
\end{table}

In the experimental results, the success rate for the Bottom-up wrapping reached 93.75\%, while the detach success rate was 95.56\%, and the overall harvesting success rate stood at 89.58\%.  It's noteworthy that successful harvesting does not solely rely on accurate prediction of the target keypoints, the precision with which the end-effector encircles the tomato trusses is equally crucial.

\subsubsection{Autonomous Continuous Harvesting Test}

In this experiment, we continued with the target selection strategy outlined in the previous section, focusing primarily on two types of target postures for harvesting. Unlike before, the environment wasn't manually pruned for this test, and the robot attempted each target only once. The robot autonomously operated in an environment teeming with potential harvesting targets, encompassing actions like: advancing, target identification, assessing harvest viability, executing the harvest, and autonomously moving to the next target location.

The main emphasis of this experiment was on: the robot's decision-making accuracy based on phenotype and posture estimation, the success rate of the harvesting operation, and its capability to execute tasks consecutively. To delve deeper, multiple tests were conducted, from which a representative experiment involving the operational records of 15 tomato trusses was selected. The specific data is presented in Table 4.

\begin{table}[htbp]
\fontsize{8}{10.8}\selectfont
\centering
\vspace{-0.5em}
\caption{Harvesting Results for Various tomato trusses in Continuous Automatic Picking Experiments in Natural Environments: 'Time Used' refers to the duration required for the robotic arm to move from its initial position to complete the cutting process.}
\label{tab:my-table}
\begin{tabular}{ccllll} 
\hline
\textbf{} & \begin{tabular}[c]{@{}c@{}}\textbf{Tomato }\\\textbf{Truss}\\\textbf{Pose}\end{tabular} & \begin{tabular}[c]{@{}l@{}}\textbf{Bottom-up }\\\textbf{Wrapping}\end{tabular} & \textbf{Detach} & \textbf{Harvesting} & \textbf{Time used (s)} \\ 
\hline
1 & Front & \multicolumn{3}{c}{{\cellcolor[rgb]{0.753,0.753,0.753}}\checkmark} & 26 \\
2 & Right & \multicolumn{3}{c}{{\cellcolor[rgb]{0.753,0.753,0.753}}\checkmark} & 21 \\
3 & Right & \multicolumn{3}{c}{{\cellcolor[rgb]{0.753,0.753,0.753}}\checkmark} & 24 \\
4 & Right & \multicolumn{3}{c}{{\cellcolor[rgb]{0.753,0.753,0.753}}\checkmark} & 22 \\
5 & Front & \multicolumn{3}{c}{{\cellcolor[rgb]{0.753,0.753,0.753}}\checkmark} & 34 \\
6 & Right & \multicolumn{1}{c}{\checkmark} & \multicolumn{1}{c}{×} & \multicolumn{1}{c}{×} & 32 \\
7 & Front & \multicolumn{3}{c}{{\cellcolor[rgb]{0.753,0.753,0.753}}\checkmark} & 27 \\
8 & Right & \multicolumn{3}{c}{{\cellcolor[rgb]{0.753,0.753,0.753}}\checkmark} & 26 \\
9 & Back & \multicolumn{1}{c}{×} & \multicolumn{1}{c}{-} & \multicolumn{1}{c}{×} & 32 \\
10 & Right & \multicolumn{3}{c}{{\cellcolor[rgb]{0.753,0.753,0.753}}\checkmark} & 28 \\
\multicolumn{1}{l}{11} & Right & \multicolumn{3}{c}{{\cellcolor[rgb]{0.753,0.753,0.753}}\checkmark} & 30 \\
\multicolumn{1}{l}{12} & Front & \multicolumn{3}{c}{{\cellcolor[rgb]{0.753,0.753,0.753}}\checkmark} & 29 \\
\multicolumn{1}{l}{13} & Right & \multicolumn{3}{c}{{\cellcolor[rgb]{0.753,0.753,0.753}}\checkmark} & 31 \\
\multicolumn{1}{l}{14} & Right & \multicolumn{3}{c}{{\cellcolor[rgb]{0.753,0.753,0.753}}\checkmark} & 30 \\
\multicolumn{1}{l}{15} & Front & \multicolumn{3}{c}{{\cellcolor[rgb]{0.753,0.753,0.753}}\checkmark} & 30 \\
\multicolumn{1}{l}{} & \multicolumn{1}{l}{} & \begin{tabular}[c]{@{}l@{}}93.33\%\\(14/15)\end{tabular} & \begin{tabular}[c]{@{}l@{}}92.86\%\\(13/14)\end{tabular} & \begin{tabular}[c]{@{}l@{}}86.67\%\\(13/15)\end{tabular} & \begin{tabular}[c]{@{}l@{}}13 successes \\in 422s,\\avg. 32.46s\end{tabular} \\
\hline
\end{tabular}
\vspace{-1.0em}
\end{table}

In the results of this experiment, we observed that the Bottom-up wrapping method based on posture estimation achieved a success rate of 93.33\%. The end effector's success rate for stem detachment stood at 92.86\%, with an overall harvesting success rate reaching 86.67\%.

Upon analyzing the factors leading to harvesting failures, we identified three primary causes: 1) Interference from the actuator leading to target displacement; 2) Errors in posture estimation; and 3) Limitations of the end effector's performance. To illustrate, the tomato trusses labeled as \#9 encountered errors in posture estimation due to occlusions, causing it to bypass the decision system's filter, and the end effector could not successfully enclose and harvest the target. Additionally, the failure in harvesting the \#6 tomato trusses occurred because the end effector rotated, with its netted bottom touching the base of the cluster, displacing it. Consequently, the stem did not fall into the guiding slot for the blade to cut, resulting in an unsuccessful harvest.

\section{Conclusion}

In this paper, we design AHPPEBot, an advanced robot specifically designed for the automated harvesting of tomatoes. To ensure the autonomy and precision of the robot's harvesting capabilities, we integrated two pivotal technologies: a rapid phenotyping method based on object detection and a pose estimation technique for tomato trusses using key points. These integrations have enhanced the robot's abilities in recognizing tomato trusses, decision-making, and planning its harvesting path. Experimental results in a greenhouse setting reveal that with effective visual guidance and a high-performing end-effector, our robot achieved state-of-the-art harvesting success rates, demonstrating its capability for continuous harvesting. The results offer a highly efficient and reliable automated solution for agricultural robotics.

\end{document}